\documentclass[10pt,letterpaper]{article}
\usepackage[T1]{fontenc}
\usepackage{times,microtype,graphicx,booktabs,tabularx,amsmath,amssymb}
\usepackage{natbib}
\usepackage[hypertexnames=false,hidelinks]{hyperref}
\usepackage{xurl}
\hypersetup{pdftitle={A Masked Future Is Not a Hidden Future: Keeping MXFP8 Attention Causal},pdfauthor={Pushkar Patel, Romit Jain, Tejesh Bhalla}}
\newcommand{\authorsymbol}[1]{\raisebox{.6ex}{\normalfont\normalsize$\scriptstyle #1$}}
\begin{document}
\twocolumn[{
\begin{center}
{\LARGE\bfseries A Masked Future Is Not a Hidden Future:\\ Keeping MXFP8 Attention Causal\par}
\vspace{1em}
{\large Pushkar Patel\authorsymbol{*} \qquad Romit Jain\authorsymbol{*\dagger} \qquad Tejesh Bhalla\authorsymbol{*}\par}
\vspace{0.35em}
Sarvam AI\par
\small
\href{mailto:pushkar@sarvam.ai}{\texttt{pushkar@sarvam.ai}} \quad
\href{mailto:romitjain@sarvam.ai}{\texttt{romitjain@sarvam.ai}} \quad
\href{mailto:tejesh@sarvam.ai}{\texttt{tejesh@sarvam.ai}}\par
\end{center}
\begin{center}\bfseries Abstract\end{center}
\begin{quotation}\small
Causal attention must prevent future tokens from influencing earlier predictions, including during parallel training and prompt prefill. Its cost grows quadratically with sequence length, making lower-precision arithmetic an attractive way to accelerate these computations. MXFP8 enables fast eight-bit matrix products on NVIDIA Blackwell GPUs, but sharing quantization scales across tokens can break causality. In Transformer Engine's stock MXFP8 attention, a masked future token can change how an earlier value is rounded and alter an earlier prediction. We demonstrate this dependency in released language models and present a causal attention kernel that removes it. Within value blocks containing both visible and future tokens, our kernel uses token-local scales and folds them into the corresponding attention weights, retaining native MXFP8 for both forward matrix products. We support training with a causal dequantized-input backward and inference with an MXFP8 KV cache. On Qwen3-4B-Base at 32K context, our kernel reduces full-sequence versus prefix disagreement on the most likely next token from 0.416\% to 0.021\%, close to the BF16 reference of 0.015\%. In end-to-end Qwen3-4B training at 32K, our recipe improves throughput by 4.9\% and reduces peak memory by 5.2\% relative to BF16. For single-request inference, it improves Qwen3-4B decode throughput by 3.8--9.9\% over vLLM with FP8 attention and an FP8 KV cache at 4K--32K context, with broadly comparable throughput on Qwen3-14B.
\end{quotation}\vspace{0.75em}}]
\begingroup
\makeatletter
\renewcommand{\thefootnote}{\ifcase\value{footnote}\or *\or \dagger\fi}
\renewcommand{\@makefnmark}{\hbox{\authorsymbol{\@thefnmark}}}
\makeatother
\footnotetext[1]{All authors contributed equally and are listed alphabetically by given name.}
\footnotetext[2]{Corresponding author: Romit Jain.}
\endgroup
\section{Introduction}

Autoregressive language models learn from complete sequences but must generate each new token using only the tokens available so far. Causal attention connects these two modes of computation. It masks future query-key scores before forming attention weights, then uses those weights to combine value vectors \citep{attention2017}. During training and prompt prefill, both matrix products are evaluated across many query positions, with work that grows quadratically with sequence length. FlashAttention reduces their memory traffic, and low-precision arithmetic can accelerate the products themselves \citep{flash2022,flash22023,flash32024,sage22024}. These optimizations must preserve the information boundary imposed by autoregressive prediction.

In MXFP8 attention, that boundary must also hold during quantization. MXFP8 stores groups of 32 elements with a shared power-of-two scale \citep{ocp2023}. In a matrix multiplication (GEMM), these groups follow the reduction dimension. For query-key scoring, each group contains features from one token. For value aggregation, each group contains the same feature from 32 consecutive tokens \citep{nvidia2026}. We call the corresponding 32 rows of V a value block. Each feature column has its own scale within that block.

\begin{figure*}[t]
\centering
\includegraphics[width=0.8\textwidth]{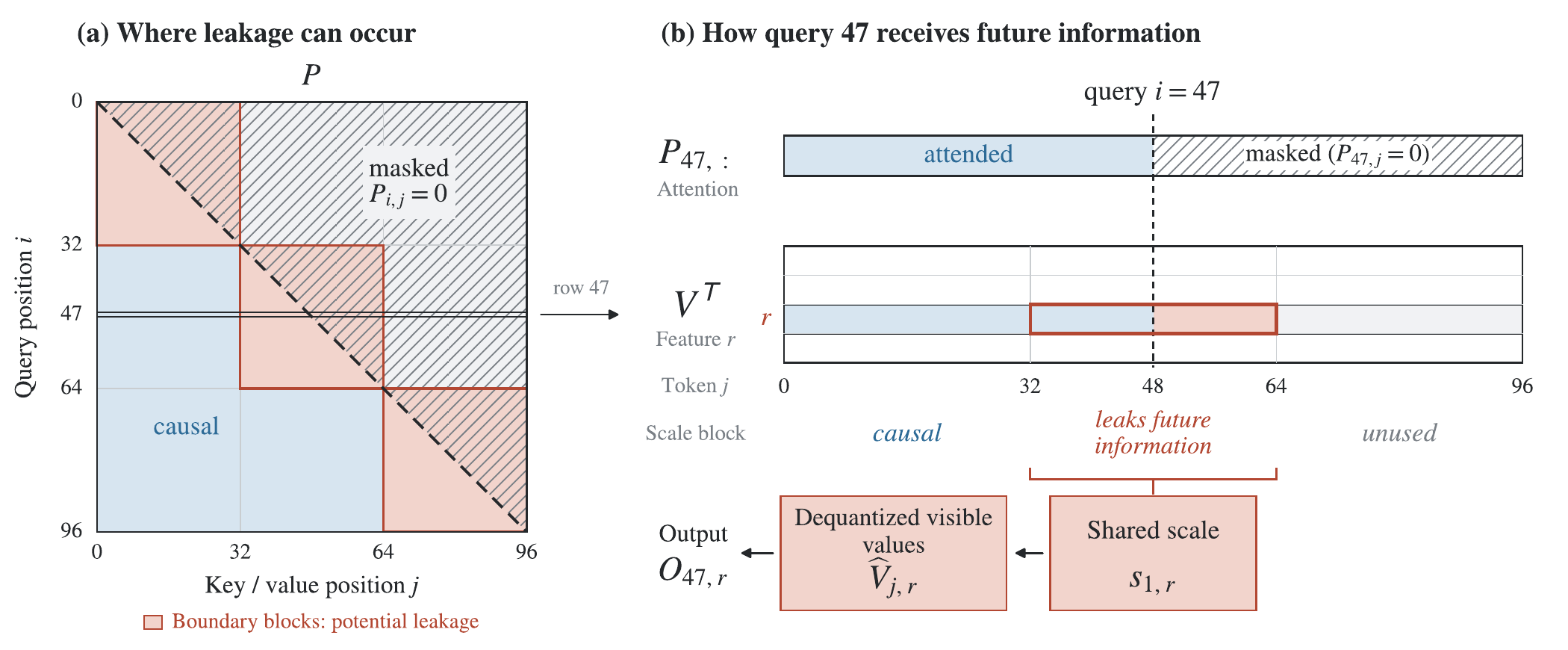}
\caption{Shared-scale leakage in MXFP8 attention.
(a) The attention matrix, shown for 96 token positions, is partitioned into $32\times32$ blocks. Blue blocks use fully visible value groups. Red diagonal blocks mark where shared scales can mix visible and future values. Hatching marks masked entries above the dashed causal diagonal, where $P_{i,j}=0$.
(b) For query $i=47$, the transposed value matrix highlights feature $r$. Tokens 32--63 share scale $s_{1,r}$, although only tokens 32--47 are visible to this query. Future values at positions 48--63 can therefore change the dequantized visible values $\widehat V_{j,r}$ and, through them, the output $O_{47,r}$. Here $\widehat V_{j,r}$ denotes the FP8 payload multiplied by its scale; it need not equal the original value $V_{j,r}$. Causal masking blocks future tokens' direct contribution but not this indirect dependency.
}
\label{fig:1}
\end{figure*}

\autoref{fig:1}(a) shows where shared value scales can introduce future dependence: within the block containing the query. Consider query 47 in \autoref{fig:1}(b). It can attend to all value rows at positions 0--31 and to rows 32--47 of the next block. Rows 48--63 are in the future. Nevertheless, the tested MXFP8 recipe in NVIDIA Transformer Engine (TE), using cuDNN fused attention, selects each scale from all 32 rows of that second value block \citep{nvidia2026,teissue2026}. We call this recipe stock MXFP8 attention. A future value at position 55 can therefore change how a visible value at position 32 is rounded. Query 47 can read this changed value even though position 55 remains masked.

MatX identified this shared-scale dependency and demonstrated its effect in MXFP4 training \citep{matx2026}. We examine MXFP8, where a power-of-two scale change often preserves a normal floating-point value but can change a value that enters the subnormal range. We give an explicit counterexample and show that the dependency affects released Qwen3 checkpoints without additional training. When we change only future values at a fixed sequence shape, stock MXFP8 changes earlier outputs while BF16 does not.

We also compare predictions made with a full sequence and with its prefix alone. Removing the suffix changes matrix shapes and floating-point reduction orders, so this comparison includes numerical effects that can occur in a causal implementation \citep{he2025nondeterminism}. Across Qwen3-4B-Base and Qwen3-14B-Base at 4K and 32K context, stock MXFP8 shows greater disagreement than BF16 and our kernel.

Our kernel retains native MXFP8 for both forward attention products, including the boundary value product. MatX computes this boundary contribution without quantization while retaining block-quantized products over earlier blocks \citep{matx2026}. We use the same partition, with token-local value scales folded into the corresponding attention weights at the boundary. Before the folded weights are rounded, this rearrangement preserves the weighted sum of the dequantized values. The resulting value and weight scales depend only on visible tokens. Combined with our dequantized-input BF16 backward, this kernel yields a training recipe with higher throughput and lower peak memory than our implementation of the MatX-style BF16-boundary approach (Section~\ref{sec:training-systems}).

We implement causal MXFP8 kernels for training and inference on NVIDIA B200 GPUs. Our contributions are threefold.

\begin{enumerate}
\item \textbf{Causal forward kernels that retain native MXFP8 arithmetic.} We introduce token-local value scaling and scale folding to remove future dependence while keeping both attention products in native MXFP8 during training and prefill.
\item \textbf{A causal training and inference recipe.} We extend the dequantized-operand approach of humans\& \citep{humans2026} from linear layers to a causal attention backward that stores no BF16 $Q/K/V$. For inference, we implement a paged MXFP8 KV cache with scales computed only from observed tokens.
\item \textbf{An analysis of causality in MXFP8 attention.} We establish how shared value scales can introduce future dependence and demonstrate its effects in released pretrained models through fixed-shape interventions and prefix-consistency evaluations. We further show that stock MXFP8 retains excess full-sequence versus prefix disagreement even after training.
\end{enumerate} 

\section{Background and Problem Formulation}
\label{sec:background}

\subsection{Autoregressive attention}

For a sequence of length $N$ and head dimension $D$, let $Q,K,V\in\mathbb{R}^{N\times D}$ denote the query, key, and value matrices of one attention head. Query position $i$ forms scores against the keys, and the causal mask excludes positions $j>i$ before the scores are normalized. The output is a weighted sum of the visible value vectors.

\begin{equation}
\begin{aligned}
S &= QK^{\mathsf T}/\sqrt D,\\
M_{i,j} &= \begin{cases}0,&j\leq i,\\-\infty,&j>i,\end{cases}\\
P &= \operatorname{softmax}_{\mathrm{row}}(S+M),\\
O &= PV,\qquad O_{i,r}=\sum_{j=0}^{i}P_{i,j}V_{j,r}.
\end{aligned}
\end{equation}

We use zero-based positions and omit batch and head indices. The first product sums over the feature dimension $D$ to produce query-key scores. The second sums over key positions to produce each output feature. cuDNN names the score and value products BMM1 and BMM2, respectively \citep{nvidia2026}.

The model uses the representation at position $i$ to predict token $i+1$. This prediction must remain independent of later tokens, whether evaluated within a full sequence or on a prefix alone.

\subsection{Scale groups in MXFP8 attention}

MXFP8 combines FP8 elements with a shared scale for every group of 32 elements. The scale uses E8M0, which represents powers of two, and the payload can use E4M3 or E5M2 \citep{ocp2023}. We use E4M3 throughout. For an element $x$ and scale $s$, quantization first divides by $s$, rounds to the FP8 format, and multiplies by $s$ when the value is used. If $\mathcal Q_{\mathrm{E4M3}}$ denotes rounding to E4M3, the represented value is

\begin{equation}
\widehat x=s\,\mathcal Q_{\mathrm{E4M3}}(x/s).
\end{equation}

MXFP8 scale groups follow the matmul's reduction dimension. In $QK^{\mathsf T}$, each group spans 32 features within one token. In $PV$, each $V$ group spans 32 consecutive key positions for one feature. A 32-token value block therefore has $D$ scales, one per feature. In the stock MXFP8 forward recipe, the probability operand ($P$) uses a fixed, data-independent scaling factor before E4M3 conversion, with the inverse factor applied in the product \citep{nvidia2026,teissue2026}. The factor depends on the kernel's softmax representation. Among these forward operands, only $V$ uses input-dependent scales shared across tokens. For query $i$, we call the block containing $i$ the \emph{boundary block}, since it may contain both visible and masked future positions. Earlier blocks are \emph{completed blocks} and are fully visible to the query.

\subsection{Prefix invariance}

Let $F_N$ denote attention at sequence length $N$, including operand quantization, and let $X=(Q,K,V)$. If two inputs agree through position $t$, changing their suffixes must leave outputs through $t$ unchanged. We call this requirement prefix invariance.

\begin{equation}
X_{0:t}=X'_{0:t}\quad\Longrightarrow\quad
F_N(X)_{0:t}=F_N(X')_{0:t}.
\end{equation}

Here $0:t$ includes position $t$. This condition applies to every operation that prepares an attention operand. A future value can influence an earlier output through scale selection even when its direct attention weight is zero. Token-local projections and normalization preserve the input prefix, so an attention violation can propagate to earlier model predictions.

We distinguish two tests throughout the paper. A fixed-shape intervention changes only the suffix while holding sequence length constant. An earlier output change then demonstrates future dependence. A full-versus-prefix comparison removes the suffix and recomputes the observed context. Its two passes predict the next token at the same position, but their different matrix shapes can also produce different rounding. The protocols, numerical controls, and results are detailed in Section~\ref{sec:checkpoints}.

\subsection{A constructive leakage example}
\label{sec:counterexample}

We now give a numerical example of the shared-scale dependency shown in \autoref{fig:1}(b). We compare two inputs that are identical except at token 55, which query 47 cannot attend to. For this example, we choose the smallest power-of-two scale $s$ such that the block's maximum absolute value divided by $s$ is at most 448, E4M3's largest finite value.

Take $N=D=128$ and $Q=K=0$. Query 47 then assigns equal probability $1/48$ to tokens 0--47 and zero probability to all later tokens. Set every value to zero except $V_{32,0}=2^{-20}$, and compare $V_{55,0}=0$ with $V_{55,0}=1$. The visible inputs and attention weights are identical in both cases. Without value quantization, both inputs therefore produce

\begin{equation}
O_{47,0}=\frac{2^{-20}}{48}.
\end{equation}

The two values belong to the same scale group, $G=\{32,\ldots,63\}$. We compare the visible value's representation before and after changing the future value.

\textbf{Before: the visible value is preserved.}
When $V_{55,0}=0$, the block maximum is $2^{-20}$. We choose the smallest power-of-two scale $s$ satisfying $2^{-20}/s\le448$, giving $s_{1,0}=2^{-28}$. The visible value enters the E4M3 quantizer as

\[
\frac{2^{-20}}{2^{-28}}=256,
\]

which is exactly representable. Multiplying the payload by the scale recovers $\widehat V_{32,0}=2^{-20}$, so its contribution to the output is preserved.

\textbf{After: the visible value rounds to zero.}
Changing $V_{55,0}$ to one raises the block maximum to one and the scale to $2^{-8}$. The unchanged visible value now enters the quantizer as

\[
\frac{2^{-20}}{2^{-8}}=2^{-12}.
\]

The smallest positive E4M3 value is $2^{-9}$. Since $2^{-12}$ is less than half that value, round-to-nearest maps it to zero \citep{ocp2023}. Thus $\widehat V_{32,0}=0$, and the only nonzero visible contribution disappears. The future value itself still has zero attention weight.

With exact attention weights, the output changes from $2^{-20}/48$ to zero even though the two inputs agree through position 47. This violates prefix invariance. The change comes entirely from how the visible value is represented. Neither its original value nor its attention weight has changed.

\subsection{The correction required at the causal boundary}

Only the value block containing query $i$ requires special treatment: earlier blocks are fully visible, and later blocks have zero attention weight \citep{matx2026}. Our objective is to remove future dependence within this boundary block while retaining native MXFP8 multiplication. Section~\ref{sec:prefill} presents our kernel, which uses token-local value scales within the boundary block and folds them into the corresponding attention weights. Section~\ref{sec:checkpoints} evaluates the kernel in released pretrained checkpoints.

\section{Causal MXFP8 Prefill}
\label{sec:prefill}

\begin{figure}[t]
\centering
\includegraphics[width=0.9\columnwidth]{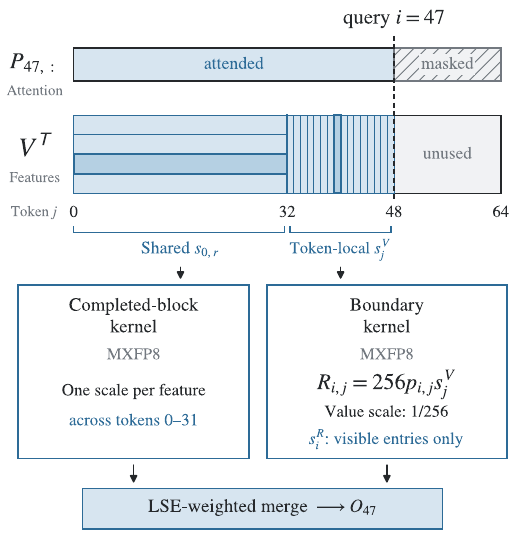}
\caption{Causal MXFP8 attention for query $i=47$. The transposed value matrix shows the scale groups used by the two kernels, which read separate value representations. The completed-block kernel shares each feature's scale $s_{0,r}$ across visible tokens 0--31. The boundary kernel instead computes a scale $s_j^V$ across the features of each token at positions 32--47 and folds it into the unnormalized attention weights $p_{i,j}$. The folded-weight scale $s_i^R$ uses only visible boundary entries; the value operand uses a fixed scale of $1/256$. Masked tokens 48--63 therefore cannot affect these scales. Darker outlines highlight one scale group of each type; feature rows are schematic.}
\label{fig:4}
\end{figure}

We compute attention over completed blocks and the visible part of the boundary block separately, as shown in \autoref{fig:4}. The \emph{completed-block kernel} reuses NVIDIA's optimized MXFP8 kernel for fully visible value blocks. The \emph{boundary kernel} uses token-local value scales and folds them into the attention weights. It retains native MXFP8 multiplication. Its weight-scale selection also uses only visible positions, preventing masked future values from influencing either representation.

For query position $i$, let $B(i)=\lfloor i/32\rfloor$. The two sets of visible keys are

\[
\begin{aligned}
\mathcal C_i&=\{j:0\le j<32B(i)\},\\
\mathcal D_i&=\{j:32B(i)\le j\le i\}.
\end{aligned}
\]

Each computation returns a partial output normalized over its own key set and the corresponding log-sum-exp (LSE). We combine these outputs using the standard LSE-based attention merge, which weights each partial output by its share of the total softmax normalization \citep{flashinfer2025}.

\subsection{Reusing the completed-block kernel}

A preparation pass constructs the MXFP8 operands from BF16 inputs using the scale groups described in Section~\ref{sec:background}. Our adaptation of NVIDIA's cuDNN Frontend SM100 MXFP8 attention kernel computes the partial result over $\mathcal C_i$ \citep{cudnnfe2026}. We restrict its attention bound from $j\le i$ to $j<32B(i)$. Queries 0--31 have no completed block and use only the boundary result.

\subsection{Folding token-local scales into the weights}

Each boundary value row receives a scale computed across its $D$ features. The boundary kernel reuses the completed-block computation's $\widehat Q$ and $\widehat K$ operands. Token-local value scaling removes the cross-token dependency in value quantization, but creates a hardware mismatch: native MXFP8 multiplication expects value scales shared across 32 tokens along the reduction dimension. We accommodate the token-dependent scales by folding them into the attention weights.

Consider two dequantized visible values, 8 and 32, with unnormalized softmax weights 1 and $1/2$. Their weighted sum is $1\cdot8+(1/2)\cdot32=24$. Both values can be represented by an FP8 payload of 256, with respective scales $8/256$ and $32/256$. Moving the factors 8 and 32 into the weights gives

\begin{equation}
1\cdot8+\frac12\cdot32
=\frac{8\cdot256+16\cdot256}{256}
=24.
\end{equation}

The value payloads now share the factor $1/256$, while the weights become 8 and 16. The softmax denominator still uses the original weights, $1+1/2=1.5$. The output therefore remains $24/1.5=16$ in this exactly representable example.

\textbf{Normalize each value row independently.}
For each nonzero row $V_j=V_{j,:}$, we select a token-local reconstruction scale from its maximum absolute feature value:

\begin{equation}
s_j^V
=2^{\left\lceil
\log_2\left(\max_r|V_{j,r}|/256\right)
\right\rceil}.
\end{equation}

We quantize and dequantize each feature using the same notation as the shared-scale conversion:

\begin{equation}
U_{j,r}
=\mathcal Q_{\mathrm{E4M3}}\!\left(
\frac{V_{j,r}}{s_j^V}\right),
\qquad
\widehat V_{j,r}=s_j^V U_{j,r}.
\end{equation}

Here $U_{j,r}$ is the FP8 payload and $\widehat V_{j,r}$ is its dequantized value. All features of token $j$ share the same scale $s_j^V$, computed only from that token. The representation is therefore independent of other tokens, and its scale can be folded into the scalar attention weight for token $j$.

Within the supported scale range, this rule bounds the scaled magnitudes by \(256=2^8\), the largest power of two below E4M3's finite limit of 448. A 448-based scale rule also performed comparably in our tests; we retain 256 as a convenient normalization choice. Causality follows from computing each scale within one token, independently of the chosen bound.

Using scores $S_{i,j}=\widehat Q_i\widehat K_j^{\mathsf T}/\sqrt D$, we compute the unnormalized softmax weights over visible boundary keys:

\[
\begin{aligned}
m_i &= \max_{j\in\mathcal D_i} S_{i,j},\\
p_{i,j} &= \exp(S_{i,j}-m_i),\\
\ell_i &= \sum_{j\in\mathcal D_i} p_{i,j}.
\end{aligned}
\]

Unlike the normalized probabilities $P$ in Section~\ref{sec:background}, $p_{i,j}$ denotes an unnormalized softmax weight.

\textbf{Move the value scales into the weights.}
For each nonzero value row, define the folded weight $R_{i,j}=256p_{i,j}s_j^V$. Each contribution can then be written as

\begin{equation}
p_{i,j}\widehat V_{j,r}
=p_{i,j}s_j^V U_{j,r}
=\frac{R_{i,j}U_{j,r}}{256}.
\label{eq:scale-folding}
\end{equation}

This identity holds for every feature $r$ with the same folded weight $R_{i,j}$. It preserves the contribution of the dequantized value exactly, before the folded weight is rounded to FP8.

In the boundary kernel shown in \autoref{fig:4}, the value payloads use the fixed scale $1/256$, while the folded weights $R_{i,j}$ carry the token-dependent factors.

As in the two-value example, the denominator remains $\ell_i$. Folding changes the numerator's operands but not its normalization. For zero value rows, we set $U_{j,r}=0$ and $R_{i,j}=0$, while retaining their original weights $p_{i,j}$ in the denominator.

\textbf{Quantize the folded weights and compute the output.}
The folded weights share a power-of-two scale selected from visible boundary entries, using E4M3's finite limit of 448.

\begin{equation}
s_i^R
=2^{\left\lceil
\log_2\left(
\max_{j\in\mathcal D_i}|R_{i,j}|/448
\right)
\right\rceil}.
\end{equation}

This expression assumes a nonzero maximum within the supported range. If all folded weights are zero, their payloads are zero and we use a unit scale. Our implementation handles scale limits explicitly.

We quantize each folded weight using the shared scale $s_i^R$:

\begin{equation}
W_{i,j}
=\mathcal Q_{\mathrm{E4M3}}\!\left(
\frac{R_{i,j}}{s_i^R}\right),
\qquad
\widehat R_{i,j}=s_i^R W_{i,j}.
\end{equation}

Here $W_{i,j}$ is the FP8 payload and $\widehat R_{i,j}$ is the dequantized folded weight. Rounding generally makes $\widehat R_{i,j}$ differ from $R_{i,j}$.

For a fixed query $i$, the scale $s_i^R$ is shared across all visible boundary keys. Substituting $\widehat R_{i,j}=s_i^R W_{i,j}$ and factoring out this scale gives the boundary output:

\begin{equation}
O_{i,r}^{\mathrm{bdry}}
=\frac{s_i^R}{256\ell_i}
\sum_{j\in\mathcal D_i}W_{i,j}U_{j,r}.
\label{eq:boundary-output}
\end{equation}

The sum multiplies the FP8 payloads $W$ and $U$. Their operand scales are $s_i^R$ and $1/256$, respectively, so the numerator can be computed with a native MXFP8 matrix product. The denominator remains $\ell_i=\sum_{j\in\mathcal D_i}p_{i,j}$, computed from the original unnormalized FP32 weights. The kernel forms folded weights in log space by adding $\log_2(256s_j^V)$ to $\log_2 p_{i,j}$ before selecting the weight scale and casting to FP8.

The boundary kernel computes $QK^\top$ and $WU$ with native block-scaled instructions. It forms the partial output $O_i^{\mathrm{bdry}}$ and its log-sum-exp (LSE), $L_i^{\mathrm{bdry}}=m_i+\log\ell_i$, then combines them with the completed-block output and LSE using the merge described above. The kernel writes the final output directly, avoiding a separate merge launch.

To organize these boundary computations, the kernel packs four query heads sharing a K/V head across 32 query positions into 128 matrix rows, reusing a single K/V tile. Other head configurations use separate packing paths. The implementation supports head dimension 128, zero dropout, and sequence lengths divisible by 128.

\section{Training and Inference Support}
\label{sec:training-support}

The prefill construction supports parallel attention in training and prompt processing. Training additionally needs a backward rule, while incremental inference needs a cache that can be updated without consulting future tokens.

\subsection{Dequantized-input BF16 backward}
\label{sec:backward}

Native MXFP8 backward computes shared quantization scales across sequence positions for $K$ in $dQ$ and $Q$ in $dK$ \citep{nvidia2026}. Future tokens can therefore alter quantized prefix operands and gradients despite causal masking. Our controls also show that replacing native backward with a BF16 surrogate leaves leakage through the saved forward output in $\delta_i=\langle G_i,O_i\rangle$, where $G$ is the incoming output gradient and $O$ is the saved forward output. Our causal forward kernel and token-local decoded operands address both paths.

Extending the dequantized-backward approach of humans\& \citep{humans2026} from linear layers to attention, we save quantized $Q/K$ and their scales, token-local $V$ payloads and amplitudes, the actual BF16 output, and FP32 log-sum-exp. Decoding operands to BF16 before backward retains their quantization error while avoiding sequence-wise requantization. Original BF16 $Q/K/V$ are not retained.

A FlashAttention-4 backward \citep{fa4} computes gradients from the decoded operands and saved forward statistics. For GQA, $dK$ and $dV$ are summed over query heads sharing a KV head. Supplement, Section~C.1 gives the surrogate equations.

This surrogate is not the exact derivative of the quantized forward: it passes gradients straight through quantization, substitutes token-local $\widehat V$ for the completed-block value representation, and omits FP8 rounding of the probabilities. It is nevertheless closer to BF16 than the stock MXFP8 backward. Across 36 synthetic cases, pooled relative L2 errors against BF16 gradients are \textbf{5.23\%}, \textbf{5.21\%}, and \textbf{3.37\%} for $dQ$, $dK$, and $dV$, compared with 6.63\%, 6.58\%, and 4.99\% for stock MXFP8. To test causality, we fix an output gradient $G$ supported only on prefix positions and multiply the suffix values $V$ by $2^{10}$. Under our backward, prefix gradients change by at most $\mathbf{6.60\times10^{-5}}$ in relative L2, within the $6.80\times10^{-5}$ maximum difference between repeated unperturbed runs. Under stock MXFP8, they change by up to $1.18\times10^{-3}$, even though its repeated unperturbed runs are bitwise identical. Deterministic BF16 gradients remain bitwise unchanged. Supplement, Sections~C.2--C.3 detail the causal dependencies and numerical checks.

\subsection{Paged decode with incremental value scaling}
\label{sec:paged-decode}

Prefill computes many query positions at once; decode receives one new token at a time. At step $t$, all tokens in the cached history precede the query and are therefore visible. We can share value scales across those tokens as long as each output uses only information available at its own step.

Our cache uses vLLM's page allocator and block table \citep{paged2023}. Each page holds 128 tokens, organized into four 32-token quantization blocks per KV head. K uses scales across 32 features within a token; V uses scales across 32 tokens within a feature. Each V block therefore has $D$ separate scales, one per feature column. A separate pass writes the prompt into the cache and computes a partial final block's scales from populated rows only.

At step $t$, the kernel first reads the cached history using the scales left by step $t-1$. The new token's self-contribution uses its BF16 key and value directly. For that single key, the normalized output is $V_t$ and the log-sum-exp is $Q_tK_t^\top/\sqrt D$. A merge kernel combines this state with the history states using the LSE-based merge described in Section~\ref{sec:prefill}.

After reading the open block, its owning program appends the new K/V. K quantization is token-local. For V, each feature retains its running scale unless the incoming value exceeds its range. A scale increase rescales at most 31 earlier payloads in the open block; completed blocks remain unchanged. No BF16 copy of earlier V rows is needed. Power-of-two rescaling preserves a dequantized value while the shifted FP8 payload remains normal; entering the subnormal range can introduce rounding or underflow. Supplement, Sections~B.3--B.4 give the update rule and causal argument.

The update becomes visible only to subsequent queries. Token 208 can change the cached representation of token 200 for query 209, which is allowed to see both tokens; it cannot change an output already produced for query 200. This ordering preserves causality. Prefill and decode nevertheless use different boundary representations, so they need not produce identical outputs for the same prefix.

The current integration supports single-token decoding and unchunked prefill with prefix caching disabled. Prompt lengths not divisible by 128 use BF16 prefill while still populating the MXFP8 cache.

\section{Leakage in Released Checkpoints}
\label{sec:checkpoints}

The leakage example in Section~\ref{sec:counterexample} establishes that shared value scales can violate prefix invariance. We now evaluate its effect in released pretrained checkpoints, using Qwen3-Base models as a case study without additional training. Each of four attention variants is compared with itself under two input conditions using the following tests.

\begin{enumerate}
\item \textbf{Fixed-shape suffix amplification.} We change only future inputs while keeping the prefix and sequence length fixed. This measures whether changing future inputs affects predictions for the unchanged prefix.
\item \textbf{Full-sequence versus prefix comparison.} We compare next-token predictions from a full-sequence pass with those from the corresponding prefix alone. This measures prediction consistency.
\end{enumerate}

Across these tests, our causal MXFP8 kernel eliminates the observed fixed-shape prefix changes and brings full-sequence versus prefix disagreement close to the BF16 reference.

\subsection{Models, variants, and data}

The models are Qwen3-4B-Base, with 36 layers, 32 query heads and eight KV heads, and Qwen3-14B-Base, with 40 layers, 40 query heads and eight KV heads \citep{qwen4,qwen14}. Both use head dimension 128 and BF16 weights. For our evaluations, only the core attention operator changes, while the remaining model operations retain their BF16 configuration. We compare Transformer Engine BF16 attention, stock MXFP8, our MXFP8 implementation of MatX's unquantized boundary correction (BF16 boundary) \citep{matx2026}, and our causal MXFP8 kernel. The two causal kernels keep MXFP8 over fully visible value blocks and differ in how they represent the boundary contribution.

The text is the English C4 validation split \citep{c42020}, tokenized with the Qwen3 tokenizer and concatenated with end-of-document tokens as in pretraining. We evaluate at context lengths of 4,096 and 32,768 tokens on NVIDIA B200 GPUs. Supplement, Sections~A.1 and~A.4 record the evaluation setup and software environment.

\subsection{Fixed-shape suffix amplification}

At a cut $c$, we multiply the value rows at positions $c$ and later by $2^{10}$ inside the attention operator of every layer. Tokens are unchanged, and $Q$ and $K$ are not perturbed directly. We test one cut per 32-token block, varying the cut's offset across successive blocks. For each cut, we compare the final logits at all positions before $c$ with the unperturbed run of the same variant, bit for bit, and measure the absolute change in next-token negative log-likelihood (NLL). Clean runs reproduce bitwise, and each reported perturbed change reproduces in a repeated run.

Multiplication by $2^{10}$ is exact in BF16 when it stays in range and is large enough to shift scales on real activations. This deliberately stresses the dependency. Supplement, Section~A.2 gives the cut schedule and numerical rationale for the factor.

We call the visible positions within the value block containing the cut \emph{at-risk positions}. Under stock MXFP8, the intervention changed the final logits at every at-risk position in both models and at both context lengths (Table~\ref{tab:suffix-amplification}). No position before the cut's value block changed. BF16, the MatX-style BF16-boundary variant, and our causal MXFP8 variant produced bitwise-identical final logits at every compared prefix position.

\begin{table}[t]
\centering
\small
\setlength{\tabcolsep}{3pt}
\caption{Fixed-shape suffix amplification under stock MXFP8 on Qwen3-Base. Counts are position--cut pairs.
The logit column reports the median per-cut maximum absolute change over prefix positions and vocabulary entries, excluding block-aligned cuts. NLL reports the mean absolute change over at-risk positions, in nats. All difference metrics are zero for BF16 and both causal kernels.}
\label{tab:suffix-amplification}
\vskip 0.1in
\begin{tabular*}{\columnwidth}{@{\extracolsep{\fill}}llrrr@{}}
\toprule
Model & Context
& \shortstack{Changed / total\\at-risk pairs}
& \shortstack{Median max.\\$|\Delta\mathrm{logit}|$}
& \shortstack{Mean\\$|\Delta\mathrm{NLL}|$} \\
\midrule
4B  & 4K  & 7,936 / 7,936   & 0.88 & 0.054 \\
    & 32K & 15,872 / 15,872 & 1.29 & 0.075 \\
\midrule
14B & 4K  & 7,936 / 7,936   & 1.13 & 0.041 \\
    & 32K & 15,872 / 15,872 & 1.41 & 0.052 \\
\bottomrule
\end{tabular*}
\end{table}

\subsection{Full-sequence versus prefix comparison}
\label{sec:prefix-comparison}

For a sampled position $c$, we run the model once on the full sequence and once on tokens $0,\ldots,c$, and compare the two next-token distributions at position $c$. The prefix pass corresponds to predicting the first token after prefilling a prompt that ends at $c$. We count a top-1 flip when the most likely next token differs between the two passes. Each model, context length, and attention variant has 163,840 comparisons. We also report the mean KL divergence between the full-sequence and prefix next-token distributions, $\mathrm{KL}(p_{\mathrm{full}}\|p_{\mathrm{prefix}})$, in nats. Supplement, Section~A.3 specifies sampling, prefix padding, and repeatability controls.

The disagreement rate for a causal kernel need not be zero. The prefix and full-sequence passes have different shapes, so matrix multiplications and attention may use different tilings and reduction orders. The resulting rounding differences can flip near-tied predictions \citep{he2025nondeterminism}. We therefore use BF16's disagreement rate as a numerical reference. Repeated full-sequence passes produced no flips in any setting.

Under stock MXFP8, the most likely next token differs in 0.39\% to 0.50\% of comparisons, versus 0.013\% to 0.059\% for BF16 and 0.013\% to 0.035\% for the two causal variants (Table~\ref{tab:prefix-consistency}). Stock MXFP8 flips 6.5 to 37 times as often as BF16. Our kernel's flip count ranges from 0.54 to 1.36 times the BF16 count across the four settings. Both causal variants reduce the excess disagreement under stock MXFP8. Both causal kernels also reduce mean KL divergence relative to stock MXFP8 in every setting, showing improved agreement across the next-token distribution.

\begin{table}[t]
\centering
\small
\setlength{\tabcolsep}{3pt}
\caption{Full-sequence versus prefix consistency on released Qwen3-Base checkpoints, with 163,840 comparisons per setting. Flips are top-1 disagreement counts; KL is mean $\mathrm{KL}(p_{\mathrm{full}}\|p_{\mathrm{prefix}})$ in nats, written in scientific notation. Lower is better. Bold and underline mark the best and second-best MXFP8 variants; BF16 is the reference.}
\label{tab:prefix-consistency}
\vskip 0.1in
\begin{tabular*}{\columnwidth}{@{\extracolsep{\fill}}lrrrr@{}}
\toprule
& \multicolumn{2}{c}{4K}
& \multicolumn{2}{c}{32K} \\
\cmidrule(lr){2-3}\cmidrule(lr){4-5}
Method & Flips & KL & Flips & KL \\
\midrule
\multicolumn{5}{@{}l}{\textbf{Qwen3-4B}} \\
BF16
& 97 & 3.15e-5
& 25 & 6.71e-6 \\
Stock MXFP8
& 634 & 3.35e-4
& 681 & 4.65e-4 \\
BF16 boundary
& \textbf{51} & \textbf{2.81e-5}
& \underline{36} & \underline{2.30e-5} \\
Ours
& \underline{58} & \underline{3.27e-5}
& \textbf{34} & \textbf{2.28e-5} \\
\midrule
\multicolumn{5}{@{}l}{\textbf{Qwen3-14B}} \\
BF16
& 71 & 2.54e-5
& 22 & 5.80e-6 \\
Stock MXFP8
& 802 & 3.61e-4
& 815 & 4.42e-4 \\
BF16 boundary
& \underline{42} & \textbf{1.71e-5}
& \textbf{22} & \textbf{1.3e-5} \\
Ours
& \textbf{38} & \underline{2.07e-5}
& \underline{23} & \underline{1.39e-5} \\
\bottomrule
\end{tabular*}
\end{table}

\section{Systems Evaluation}
\label{sec:systems}

We evaluate the end-to-end throughput and memory cost of training with causal MXFP8 attention on NVIDIA B200 GPUs, using BF16 attention as the baseline. We replace only the core attention implementation and keep the rest of the training setup fixed. We also evaluate inference throughput and numerical agreement against vLLM configurations using BF16 attention and fixed per-tensor FP8 scaling.

\subsection{Training throughput and memory}
\label{sec:training-systems}

We compare four attention implementations in Megatron-LM \citep{megatron2019}: BF16, stock MXFP8, the MatX-style BF16-boundary correction, and our complete recipe. Projections, normalization, positional embeddings, and MLPs retain the baseline implementation and precision, with all linear layers remaining BF16. Within each context length, all variants use the same initialization, data order, optimizer, batch size, and activation-recomputation settings.

The BF16 baseline uses Transformer Engine with automatic attention-backend selection. Stock MXFP8 uses the shipped cuDNN backward, the BF16-boundary variant uses FlashAttention-4 backward on the original BF16 $Q/K/V$, and ours uses the dequantized-input BF16 surrogate described in Section~\ref{sec:backward}.

Each Qwen3-4B run trains for approximately 5B C4 tokens on 16 B200 GPUs, with $2^{21}$ tokens per optimization step. At 4K, the microbatch contains four sequences per GPU and uses no activation recomputation. At 32K, it contains one sequence per GPU and selectively recomputes MLPs and normalization layers. Table~\ref{tab:training-systems} reports complete-step training results alongside standalone attention timings. Attention microbenchmarks use synthetic BF16 inputs with matching microbatch sizes and layouts ($H_q=32$, $H_{kv}=8$, $D=128$), reporting medians over six rounds of 20 CUDA-graph replays with alternating configuration order. Supplement, Sections~D.1--D.2 give the training configuration and measurement protocol.

\begin{table*}[t]
\centering
\small
\caption{Qwen3-4B systems evaluation. Attention latencies are measured on one B200, including preparation and backward preprocessing. Forward, backward, and whole attention are timed separately; their measurements need not be additive. Full-training throughput and peak memory are measured on 16 B200 GPUs. Bold and underline mark the best and second-best MXFP8 recipes; BF16 is the reference. BF16 boundary denotes the MatX-style correction. Stock MXFP8 exhibits future-token dependence.}
\label{tab:training-systems}
\vskip 0.1in
\begin{tabular}{@{}llrrrrr@{}}
\toprule
& & \multicolumn{3}{c}{Attention latency (ms)}
& \multicolumn{2}{c}{Full training} \\
\cmidrule(lr){3-5}\cmidrule(lr){6-7}
Context & Method
& Forward & Backward & Whole
& Tokens/s/GPU & Peak GiB \\
\midrule
4K & BF16
& 0.461 & 1.376 & 2.059
& 33,398 & 110.0 \\
& Stock MXFP8
& \textbf{0.612} & 1.769 & \textbf{2.213}
& \textbf{33,056} & \textbf{110.3} \\
& BF16 boundary
& 0.760 & \underline{1.542} & 2.430
& 32,258 & 120.2 \\
& Ours
& \underline{0.744} & \textbf{1.448} & \underline{2.298}
& \underline{32,702} & \underline{111.3} \\
\midrule
32K & BF16
& 6.625 & 19.644 & 26.279
& 16,805 & 124.9 \\
& Stock MXFP8
& \textbf{5.776} & \underline{18.923} & \underline{24.591}
& \underline{17,313} & \underline{125.4} \\
& BF16 boundary
& \underline{6.020} & 19.152 & 25.054
& 17,111 & 136.2 \\
& Ours
& 6.059 & \textbf{17.775} & \textbf{23.714}
& \textbf{17,624} & \textbf{118.4} \\
\bottomrule
\end{tabular}
\end{table*}

At 4K, preparation costs are less easily amortized: even stock MXFP8 trains 1.0\% slower than BF16, while ours is 2.1\% slower and uses 1.2\% more peak memory. Our preparation, completed-block attention, and boundary-plus-merge stages take 0.069, 0.506, and 0.217 ms at 4K, versus 0.127, 5.528, and 0.391 ms at 32K. These independently measured stage times are not additive.

At 32K, our attention forward and backward latencies are \textbf{8.5\% and 9.5\% lower than cuDNN BF16}, with 9.8\% lower whole-attention latency. Our training recipe delivers \textbf{4.9\% higher throughput} than BF16 while reducing peak memory by \textbf{5.2\%}, from 124.9 to 118.4 GiB. It also exceeds stock MXFP8 throughput by 1.8\%, while removing the future dependence demonstrated in Section~\ref{sec:checkpoints}.

Compared with our implementation of the MatX-style BF16-boundary recipe, ours increases end-to-end training throughput by 1.4\% at 4K and 3.0\% at 32K, while reducing peak memory by 7.4\% and 13.1\%, respectively. Both causal kernels produce bitwise-unchanged prefix logits under the fixed-shape suffix interventions in Section~\ref{sec:checkpoints}. Our kernel achieves this while retaining native MXFP8 for both forward products, including the boundary contribution. These efficiency gains reflect the complete training recipe, including the dequantized-input BF16 backward and saved-state representation; the 32K forward alone is 0.6\% slower than the BF16-boundary forward.

\subsection{Inference throughput and prefill fidelity}
\label{sec:inference-systems}

We evaluate the released \texttt{Qwen/Qwen3-4B-Base} and \texttt{Qwen/Qwen3-14B-Base} checkpoints on one NVIDIA B200 using vLLM 0.31.0, with BF16 weights and projections. Table~\ref{tab:3} compares our MXFP8 implementation with three vLLM configurations: BF16 attention with BF16 K/V storage, BF16 attention with FP8 K/V storage, and FP8 attention with FP8 K/V storage. The FP8 configurations use fixed per-tensor scales of 1.0, following the uncalibrated configuration evaluated by vLLM \citep{kubler2026fp8kvcache}; our implementation uses data-dependent block scales restricted to observed tokens.

Measurements use CUDA graphs, one request at a time, repeated-sentence prompts of 4K--32K tokens, and 128 generated tokens. Each reported value averages two rounds' medians of five requests, with reversed configuration order. Differences below approximately 3\% are comparable to observed run-to-run variation. Supplement, Section~D.3 describes the serving protocol.

\begin{table}[t]
\centering
\small
\setlength{\tabcolsep}{3pt}
\caption{Single-request decode throughput (tokens/s) on one B200, with BF16 weights and projections. Baseline labels give vLLM attention/KV-cache precision. Gain is our throughput increase over FP8/FP8.}
\label{tab:3}
\vskip 0.1in
\begin{tabular*}{\columnwidth}{@{\extracolsep{\fill}}lrrrr@{}}
\toprule
Attention/cache & 4K & 8K & 16K & 32K \\
\midrule
\multicolumn{5}{@{}l}{\textbf{Qwen3-4B}} \\
BF16/BF16 & 345.7 & 318.3 & 301.3 & 274.8 \\
BF16/FP8  & 345.7 & 312.8 & 298.5 & 272.1 \\
FP8/FP8   & 331.0 & 327.3 & 315.0 & 300.7 \\
Ours      & 363.7 & 345.5 & 334.0 & 312.2 \\
Gain      & +9.9\% & +5.6\% & +6.0\% & +3.8\% \\
\midrule
\multicolumn{5}{@{}l}{\textbf{Qwen3-14B}} \\
BF16/BF16 & 163.6 & 162.6 & 154.9 & 145.8 \\
BF16/FP8  & 158.7 & 158.6 & 150.9 & 144.3 \\
FP8/FP8   & 164.1 & 161.3 & 158.0 & 153.3 \\
Ours      & 164.9 & 165.4 & 164.7 & 155.6 \\
Gain      & +0.5\% & +2.5\% & +4.2\% & +1.5\% \\
\bottomrule
\end{tabular*}
\end{table}

Our implementation delivers \textbf{3.8--9.9\% higher decode throughput} than vLLM's FP8-attention path on Qwen3-4B. On Qwen3-14B, throughput is broadly comparable, with measured increases of 0.5--4.2\%.
Against BF16 attention with FP8 K/V storage, the increases are 5.2--14.7\% and 3.9--9.1\%, respectively. These measurements compare complete inference implementations.

We also measure prefill NLL on 16 C4 sequences of 8,192 tokens per model
(131,056 scored labels). Relative to vLLM's FP8-attention path, our NLL
decreases by $0.0035 \pm 0.0012$ on 4B and $0.0016 \pm 0.0004$ on 14B,
where uncertainty denotes one paired standard error across sequences
(Supplement, Section~D.4).
NLL is comparable to BF16 attention with FP8 K/V storage.
These prefill results do not measure incremental value-rescaling error
during decode.

\section{Prefix Consistency after Training}
\label{sec:trained-dependence}

We finally ask whether training with stock MXFP8 reduces the excess disagreement between full-sequence and prefix predictions. To test this, we train Qwen3-4B from random initialization for 5B C4 tokens at 4K and 32K context using stock MXFP8 and compare it with matched runs using BF16, the BF16-boundary correction, and our causal MXFP8 attention. The runs use the matched training setup in Section~\ref{sec:training-systems}. We repeat the full-sequence versus prefix comparison from Section~\ref{sec:prefix-comparison} on the final checkpoints, using each model's training attention variant and the protocol in Supplement, Section~A.3. At each sampled position, we compare the next-token prediction from the full-sequence pass with that from prefilling only the prefix ending at that position. Table~\ref{tab:trained-prefix-consistency} reports validation loss and top-1 flip counts, with 163,840 comparisons per setting.

\begin{table}[t]
\centering
\small
\setlength{\tabcolsep}{3pt}
\caption{Qwen3-4B after 5B training tokens. Loss uses each checkpoint's training attention variant; flips count full-sequence versus prefix top-1 disagreements across 163,840 comparisons. Lower is better. Bold and underline mark the best and second-best MXFP8 recipes respectively with BF16 as the reference.}
\label{tab:trained-prefix-consistency}
\vskip 0.1in
\begin{tabular*}{\columnwidth}{@{\extracolsep{\fill}}lrrrr@{}}
\toprule
& \multicolumn{2}{c}{4K} & \multicolumn{2}{c}{32K} \\
\cmidrule(lr){2-3}\cmidrule(lr){4-5}
Method & Loss & Flips & Loss & Flips \\
\midrule
BF16
& 3.0137 & 46
& 3.0756 & 15 \\
Stock MXFP8
& \underline{3.0150} & 123
& 3.0774 & 64 \\
BF16 boundary
& 3.0166 & \textbf{25}
& \textbf{3.0682} & \underline{19} \\
Ours
& \textbf{3.0129} & \underline{29}
& \underline{3.0722} & \textbf{18} \\
\bottomrule
\end{tabular*}
\end{table}

Stock MXFP8 continues to exhibit excess disagreement, with 4.2 times as many flips as our implementation at 4K and 3.6 times as many at 32K. Although the disagreement rates are lower than in the released checkpoints, training does not eliminate stock MXFP8's excess full-sequence versus prefix disagreement in these runs.

\section{Related Work and Implications}
\label{sec:related-work}

MatX described future-token dependence in block-quantized attention and removed it by computing the block-diagonal $PV$ contribution without quantization \citep{matx2026}. Its model experiment uses MXFP4. A public Transformer Engine issue reports the shared-scale problem in MXFP8 attention \citep{teissue2026}. A draft FlashAttention implementation similarly uses BF16 probabilities and original BF16 values for blocks intersecting the causal diagonal, with suffix-perturbation tests \citep{fa3022}. Our contribution retains native MXFP8 for both forward attention products, including the causal boundary, and evaluates future dependence in released checkpoints and prefix consistency after training.

Low-precision attention kernels make different precision and scaling choices for the two products. FlashAttention-3 uses FP8 tensor cores on Hopper with tile-level scaling and orthogonal transforms \citep{flash32024}. SageAttention retains FP16 value aggregation, and SageAttention2 uses FP8 value aggregation with per-channel value scales computed across the sequence \citep{sage2024,sage22024}. SageAttention3 also explores 8-bit forward and backward attention for training \citep{sage32025}. VC-Attention groups value tokens for low-bit non-causal video attention \citep{vc2026}. These approaches motivate examining the information used during quantization, alongside numerical error against a high-precision reference.

Attn-QAT and Delta-Matching study numerical consistency between low-precision forward and backward computations, including the softmax-gradient reduction \citep{attnqat2026,deltamatching2026}. Preserving causal dependencies is a separate requirement; our dequantized-input backward retains the approximations described in Section~\ref{sec:backward}.

KV-cache quantization raises a related question about when groups become available. KIVI quantizes keys per channel and values per token, retaining recent data in a high-precision residual cache \citep{kivi2024}. KVQuant calibrates non-uniform representations and per-channel key scaling parameters offline, while computing value scaling parameters online per token \citep{kvquant2024}. Our cache maintains running value scales within its open block, with updates visible only to subsequent queries. Its page allocation follows PagedAttention, and its parallel processing of cached history follows Flash-Decoding \citep{paged2023,flashdecoding2023}.

These findings are relevant to systems that generate tokens sequentially and score them in parallel, including RL training. Batch-dependent reduction orders can already produce numerical disagreement between these computations \citep{he2025nondeterminism}. Future-dependent quantization introduces an additional input dependency. Evaluating its downstream effects remains future work. Removing this dependency does not guarantee bitwise agreement across kernels, shapes, or cache representations.

\section{Conclusion}

Causal masking does not suffice when quantization scales depend on future tokens. Our counterexample and fixed-shape checkpoint tests establish this dependency in stock MXFP8 attention, while prefix comparisons show excess prediction disagreement.

Our central contribution is a causal MXFP8 attention kernel that removes this dependency through token-local value scaling and scale folding while retaining native MXFP8 for both forward attention products, including the causal boundary. Our causal dequantized-input BF16 surrogate backward enables training with this causal forward, and a paged MXFP8 KV cache supports inference. In end-to-end Qwen3-4B training at 32K context, our recipe delivers 3.0\% higher throughput and uses 13.1\% less peak memory than our implementation of the MatX-style BF16-boundary recipe. Relative to BF16, it delivers 4.9\% higher throughput and uses 5.2\% less peak memory. For single-request Qwen3-4B decoding on one B200 at 4K--32K context, it improves throughput by 3.8--9.9\% over vLLM with FP8 attention and an FP8 KV cache. Together, these kernels remove future-token leakage while delivering measured training and inference benefits.

\section*{AI Usage Disclosure}

OpenAI Codex and Anthropic Claude Code assisted with literature and documentation searches, discussion of the experimental protocol, drafting and revising this manuscript, LaTeX conversion, figure-generation code, and the training and evaluation scripts. The kernels were optimized in part through AI-assisted optimization loops with recorded correctness gates and experiment logs. The authors reviewed all prose, derivations, citations, code, figures and measurements and are responsible for them. AI systems are not authors.
\clearpage
\section*{Supplementary Material}
\appendix
\numberwithin{table}{section}
\numberwithin{equation}{section}
\renewcommand{\thetable}{\Alph{section}\arabic{table}}
\renewcommand{\theequation}{\Alph{section}\arabic{equation}}
This appendix provides the checkpoint-evaluation protocols, incremental cache design, surrogate-backward equations and checks, and systems measurement methodology supporting the main paper.

\section{Evaluation protocols}
\label{app:protocols}

\subsection{Data and attention variants}
\label{app:checkpoint-setup}

The released-checkpoint experiments use Qwen3-4B-Base and Qwen3-14B-Base with BF16 weights, head dimension 128, and English C4 validation text tokenized with the Qwen3 tokenizer. Documents are concatenated with end-of-document tokens. The four variants are Transformer Engine BF16 attention, stock MXFP8 through its cuDNN fused-attention backend, our MXFP8 implementation of MatX's BF16-boundary correction, and our causal MXFP8 attention. Operations outside core attention retain the baseline configuration.

\subsection{Fixed-shape suffix amplification}
\label{app:suffix-amplification}

For each cut $c$, the intervention multiplies value rows at positions $c$ and later by $2^{10}$ inside every layer's attention operator. The intervention acts only on suffix V rows; the token sequence and sequence length stay fixed, and Q and K are not perturbed directly. We compare final logits at all positions before $c$ with an unperturbed pass of the same variant and measure absolute changes in next-token NLL.

For a 32-token value block with zero-based index $b$, the cut is

\begin{equation}
\label{eq:supp-1}
c_b=32b+((7b+1)\bmod 32).
\end{equation}

Because 7 and 32 are coprime, the offsets cycle through all 32 positions over successive blocks. Each context window contributes one cut per block. Four 4K windows give 512 cuts, of which 496 have at-risk rows; one 32K window gives 1,024 cuts, of which 992 have at-risk rows. The at-risk positions run from the start of the cut's value block through $c-1$. They contribute 7,936 position--cut pairs at 4K and 15,872 at 32K. The full prefixes contain 1,048,320 and 16,776,704 position--cut pairs, respectively. Cuts on block boundaries provide controls with no at-risk rows.

At 4K, eight cuts share a batched forward; at 32K, cuts are evaluated individually. Each clean comparison uses the same batch shape as its perturbed pass. The clean pass is repeated to check bitwise reproducibility. Any perturbed change is checked in a repeated pass. The recorded runs have no discarded unstable cases. Reported logit and NLL changes concern the final model output.

Multiplication by $2^{10}$ changes the BF16 exponent exactly when the result remains in range. It provides a strong intervention on scale selection without adding rounding to the amplified rows. A larger shared scale can move smaller visible values into E4M3's subnormal range or round them to zero. The intervention deliberately stresses the shared-scale dependency; its measured magnitudes do not estimate how often ordinary text triggers that dependency.

\subsection{Full-sequence versus prefix-prefill comparison}
\label{app:prefix-comparison}

For a sampled position $c$, we compare the next-token distribution from a full-sequence pass at $c$ with the distribution obtained by prefilling only tokens $0,\ldots,c$. Both predict token $c+1$ from the same observed prefix. This is a separate prefix-prefill pass, not a cached-decode rollout.

At 4K, 20,480 sequences contribute eight sampled positions each. At 32K, 5,120 sequences contribute 32 positions each. Both settings therefore contain 163,840 comparisons per model and attention variant. Using seed 1234, we sample positions uniformly with replacement from $0,\ldots,S-2$, where $S$ is the sequence length. Each microbatch draws its own positions, with the same draws on all eight data-parallel GPUs. Microbatches contain four sequences per GPU at 4K and one at 32K, so a sampled position is shared by 32 or eight sequences, respectively. All variants use the same text and sampled positions within each context length.

The attention wrapper pads prefixes with zero Q/K/V rows to a multiple of 1,024 for stock MXFP8 and both causal variants, then removes the padded outputs. BF16 uses Transformer Engine's automatic backend selection. Appended zero rows are causally masked and cannot increase a value group's maximum. Different full and prefix execution shapes can nevertheless change tilings, reduction orders, and rounding. BF16 therefore serves as a numerical reference, rather than a requirement of zero disagreement \citep{he2025nondeterminism}.

We count a top-1 flip when the most likely next token differs. For released checkpoints, we also report mean $\mathrm{KL}(p_{\mathrm{full}}\|p_{\mathrm{prefix}})$ in nats, computed from FP32 log-softmax values. Repeated full-sequence passes provide a repeatability control and produced no top-1 flips in the reported settings. The comparisons share sequences and sampled positions, so the reported counts and ratios are descriptive rather than independent observations for significance testing.

The final trained checkpoints use this same full-sequence versus prefix-prefill protocol with each checkpoint's training attention variant. The paper reports their top-1 flip counts.

\subsection{Recorded evaluation environment}
\label{app:evaluation-environment}

The fixed-shape intervention records identify NVIDIA B200, PyTorch \texttt{2.13.0a0+8145d630e8.nv26.06}, CUDA 13.3, Transformer Engine \texttt{2.18.0+15288857}, cuDNN 9.21.1, and Transformers 5.12.1. These versions are specific to that experiment. The full-sequence versus prefix experiments run through Megatron-LM \citep{megatron2019} using checkpoints converted from the released Hugging Face models, or the final checkpoints from the matched training runs. The prefix-comparison results and source are retained in the training-results snapshot at revision \texttt{6282b6f}; the intervention records are at revision \texttt{11c3260} and include source hashes. The two evaluation suites are recorded separately; the software versions above apply specifically to the fixed-shape intervention.

\section{Incremental MXFP8 cache}
\label{app:cache}

\subsection{Page layout}
\label{app:cache-layout}

Table~\ref{tab:supp-cache-layout} shows the page layout. The cache uses vLLM's page allocator and block table \citep{paged2023}. For head dimension 128, each page holds 128 tokens per KV head, organized as four 32-token quantization blocks. K scales span 32 features within a token; V scales span 32 tokens within one feature. V is stored transposed for the value product, so its quantization groups follow the token reduction axis \citep{ocp2023}.

\begin{table*}[t]
\centering
\small
\caption{Byte layout of a cache page for one KV head with head dimension 128. Each row contains 128 bytes.}
\label{tab:supp-cache-layout}
\begin{tabularx}{\textwidth}{@{}l l X@{}}
\toprule
Page rows (128 bytes each) & Contents & Layout or scale granularity \\
\midrule
0--127 & K payloads, E4M3 & Token by feature \\
128--255 & V payloads, E4M3 & Feature by token \\
256--259 & K scales, E8M0 & Four scales per token \\
260--263 & V scales, E8M0 & 128 scales per 32-token block \\
\bottomrule
\end{tabularx}
\end{table*}

Payloads and scales occupy $128+128+4+4=264$ bytes per token per KV head, compared with 512 bytes for BF16 K/V. The ratio $512/264=1.94$ describes capacity for the same KV-cache storage budget; it is not a reduction in total model memory. A separate pass writes the prompt into the cache, uses only populated rows to scale a partial final block, and clears unused slots.

\subsection{Read, self-contribution, and append}
\label{app:cache-ordering}

At step $t$, the cache contains tokens $0,\ldots,t-1$, and the new inputs are BF16 $Q_t,K_t,V_t$. Parallel programs read partitions of the cached history and produce partial attention states, following Flash-Decoding \citep{flashdecoding2023}. The open block is read using the scales left by step $t-1$.

The new token's self-contribution uses its BF16 key and value directly. Its normalized output and log-sum-exp are

\begin{equation}
\label{eq:supp-2}
(O_{\mathrm{self}},L_{\mathrm{self}})
=\left(V_t,\;Q_tK_t^{\mathsf T}/\sqrt D\right).
\end{equation}

The LSE-based merge described in the main paper's Section~3 combines this state with the cached-history states. Once the program responsible for the open block finishes reading its old contents, it quantizes and stores the new K/V. The append is fused into the attention launch; the partial-state merge is a separate launch. K quantization is token-local and cannot change earlier K rows.

\subsection{Updating the open value block}
\label{app:cache-update}

For feature $f$, let $s_f$ be its current scale and $x_{j,f}$ an earlier token's stored E4M3 payload, representing $s_fx_{j,f}$. For a positive magnitude within the supported scale range, write

\begin{equation}
\label{eq:supp-3}
E(x)=2^{\lceil\log_2(x/448)\rceil}
\end{equation}

for the smallest power-of-two scale that accommodates it. Let $\mathcal Q_8$ denote E4M3 rounding. Appending $v_{t,f}$ applies

\begin{equation}
\label{eq:supp-4}
\begin{aligned}
s'_f&=\begin{cases}
s_f,&|v_{t,f}|\le448s_f,\\
E(|v_{t,f}|),&|v_{t,f}|>448s_f,
\end{cases}\\
x'_{j,f}&=\mathcal Q_8\!\left(x_{j,f}\frac{s_f}{s'_f}\right),\\
x_{t,f}&=\mathcal Q_8(v_{t,f}/s'_f).
\end{aligned}
\end{equation}

If the value fits, only its new payload is stored. Otherwise, the scale grows by a power of two and earlier payloads in that feature are divided by the same factor. At most 31 earlier tokens are touched; completed blocks remain unchanged. The update uses stored FP8 payloads and needs no BF16 copy of earlier values.

The identity $s'_f(x_{j,f}s_f/s'_f)=s_fx_{j,f}$ preserves the dequantized value before rounding. In E4M3, it remains exact while the shifted payload is normal; entering the subnormal range can introduce rounding or underflow. For example, at scale $1/16$, token 200's payload 320 represents 20. If token 208 brings value 50, the required payload 800 exceeds 448. Raising the scale to $1/8$ gives the new token payload 400 and changes token 200's payload to 160, still representing 20. Query 208 reads the previous cache state and its BF16 self-term; subsequent queries read the updated state.

A new 32-token block initializes its scales from its first token and clears unused slots. When the block fills, its scales become final. Zero values never raise an established scale; all-zero features use zero payloads and the implementation's designated scale encoding.

\subsection{Causal argument and scope}
\label{app:cache-causality}

At the start of step $t$, every cached payload and scale depends only on tokens before $t$. The output reads that state and its own BF16 inputs. The append depends only on the previous cache and token $t$, so the state seen by query $t+1$ depends only on tokens through $t$. This ordering holds across block and page boundaries. A new token can change an earlier token's stored representation only for subsequent queries that are allowed to observe the new token; it cannot change an output already produced.

Prefill uses token-local values at each query's boundary, whereas decode uses running cache scales and a BF16 self-term. Both paths respect the causal constraint, but their different representations need not produce identical outputs. The evaluated plugin supports single-token decoding and unchunked prefill with prefix caching disabled. Prompt lengths not divisible by 128 use BF16 prefill while still populating the MXFP8 cache.
\section{Dequantized-input BF16 backward}
\label{app:backward}

\subsection{Saved state and surrogate equations}
\label{app:backward-equations}

Extending the dequantized-backward approach of humans\& \citep{humans2026} from linear layers to attention, the forward saves quantized $Q/K$ and their scales, token-local $V$ payloads and amplitudes, the actual BF16 output $O$, and FP32 log-sum-exp $L$. Original BF16 $Q/K/V$ are not retained. Before backward, the saved payloads are dequantized into temporary BF16 operands $\widehat Q$, $\widehat K$, and $\widehat V$. The token-local $\widehat V$ can differ from the completed-block value representation used during forward.

Let $G$ be the incoming output gradient. The FlashAttention-4 backward \citep{fa4} computes the following surrogate, with $\delta$ broadcast across keys.

\begin{equation}
\label{eq:supp-backward}
\begin{aligned}
\widetilde P_{ij}
&=
\begin{cases}
\exp(\widehat Q_i\widehat K_j^{\mathsf T}/\sqrt D-L_i), & j\le i,\\
0, & j>i,
\end{cases}\\
\delta_i &= \sum_r G_{ir}O_{ir},\\
dS &= \widetilde P\odot(G\widehat V^{\mathsf T}-\delta),\\
dQ &= dS\widehat K/\sqrt D,\\
dK &= dS^{\mathsf T}\widehat Q/\sqrt D,\\
dV &= \widetilde P^{\mathsf T}G.
\end{aligned}
\end{equation}

For GQA, $dK$ and $dV$ are summed over query heads sharing the same KV head. The implementation forms the products in tiles rather than materializing a quadratic probability matrix.

These equations pass gradients through quantization using a straight-through approximation. They recompute probabilities without FP8 weight rounding, substitute token-local $\widehat V$ for the completed-block value representation, and use the actual forward output in $\delta$. Consequently, they are not the exact derivative of the rounded forward computation. Decoding preserves the quantization error already present in the saved operands; it avoids an additional quantization across sequence positions during backward.

\subsection{Causal dependencies}
\label{app:backward-causality}

Consider a fixed upstream gradient supported only before cut $c$, so $G_i=0$ for $i\ge c$. The dequantized operands at positions before $c$ depend only on those tokens, and the causal forward produces prefix outputs $O_i$ and statistics $L_i$ independent of the suffix. Equation~\ref{eq:supp-backward} therefore uses unchanged operands for every nonzero prefix contribution. Rows of $dS$ at $i\ge c$ vanish because both $G_i$ and $\delta_i$ are zero. For $i<c$, masking excludes keys $j\ge c$. Thus prefix gradients are independent of the suffix in the declared surrogate, and gradients with respect to suffix inputs are zero. Floating-point implementation checks compare prefix-gradient changes with repeated-run variation.

Both the saved forward state and the backward operands matter. Replacing native backward with BF16 products does not repair an output $O_i$ that already depends on future values: that dependency enters $\delta_i=\langle G_i,O_i\rangle$. Conversely, treating quantization scales as constants can produce zero suffix gradients while leaving prefix gradients dependent on a suffix-dependent quantized operand. Zero suffix gradients alone therefore do not establish prefix invariance.

\subsection{Numerical agreement and suffix interventions}
\label{app:backward-checks}

Across the 36 synthetic cases reported in the main paper, the surrogate has lower pooled relative L2 error against BF16 gradients than stock MXFP8 for each of $dQ$, $dK$, and $dV$ (Table~\ref{tab:supp-backward-errors}). This comparison measures error relative to BF16 attention, rather than an exact derivative of the quantized forward.

\begin{table}[t]
\centering
\small
\caption{Pooled gradient relative L2 errors against BF16 over 36 synthetic cases.}
\label{tab:supp-backward-errors}
\begin{tabular}{@{}lrrr@{}}
\toprule
Method & $dQ$ & $dK$ & $dV$ \\
\midrule
Stock MXFP8 & 6.63\% & 6.58\% & 4.99\% \\
Ours & 5.23\% & 5.21\% & 3.37\% \\
\bottomrule
\end{tabular}
\end{table}

The suffix intervention fixes $G$ with support only on prefix positions, holds the tensor shape and prefix inputs fixed, and multiplies suffix $V$ rows by $2^{10}$. Prefix gradients are compared with an unperturbed run, and repeated unperturbed runs provide a numerical control. Under our implementation, the maximum relative L2 change in prefix gradients is $6.60\times10^{-5}$; the maximum difference between repeated unperturbed runs is $6.80\times10^{-5}$. Under stock MXFP8, the intervention changes prefix gradients by up to $1.18\times10^{-3}$, while its repeated unperturbed runs are bitwise identical. Deterministic BF16 gradients remain bitwise unchanged.

These maxima summarize the tested cases; they do not establish bitwise invariance of our backward. The intervention tests future-value dependence in the combined forward and backward computation. Because it perturbs $V$, it does not by itself isolate the additional sequence-wise $Q/K$ quantization in native MXFP8 backward. The separate control using BF16 backward with the leaky forward identifies a dependency through the saved forward output. Model-training results evaluate the resulting recipe separately from these synthetic checks.

\section{Systems measurement protocols}
\label{app:systems-protocols}

\subsection{Matched training configuration}
\label{app:training-configuration}

The training study uses the Qwen3-4B architecture with 36 layers, hidden size 2,560, MLP width 9,728, 32 query heads, eight KV heads, and head dimension 128. It uses Q/K normalization, RMSNorm with epsilon $10^{-6}$, SwiGLU, tied input/output embeddings, and rotary base $10^6$. Attention and hidden dropout are zero. All runs begin from random initialization with seed 1234; operations outside core attention retain the same BF16 configuration.

Data are the English C4 corpus \citep{c42020}, tokenized with the Qwen3 tokenizer. Preparation uses 48 training shards and all eight validation shards. Documents are concatenated with end-of-document tokens and packed into full sequences without an additional cross-document attention mask. Within each context length, the variants share the sample-order cache, data order, and initialization.

Each run uses two nodes with eight B200 GPUs each, with data parallelism and a distributed optimizer. Tensor, pipeline, and context parallelism are disabled. Gradient reduction and parameter gathering overlap with computation. Every optimization step processes $2^{21}$ tokens. At 4K, the global batch is 512 sequences and the microbatch is four sequences per GPU; at 32K, these are 64 and one, respectively. The 4K runs use no activation recomputation. The 32K runs selectively recompute MLPs and normalization layers, without recomputing attention.

Training lasts 2,384 optimization steps, or 4,999,610,368 tokens. Adam uses $(\beta_1,\beta_2)=(0.9,0.95)$, epsilon $10^{-8}$, weight decay 0.033, and gradient clipping at 1.0. The learning rate warms up for 238 steps to $3\times10^{-4}$ and then follows a cosine decay to $3\times10^{-5}$. The runs use Megatron-LM in the NeMo 26.08 container.

The BF16 recipe uses Transformer Engine's automatic attention-backend selection. Stock MXFP8 uses the shipped cuDNN forward and backward. The BF16-boundary recipe uses our implementation of the unquantized boundary product and a FlashAttention-4 backward on original BF16 $Q/K/V$. Our recipe uses native MXFP8 forward products and the dequantized-input BF16 backward in Section~\ref{app:backward}.

Final validation loss is evaluated at checkpoint 2,384 using each run's training attention variant over 80 global validation batches, approximately 168 million tokens. The full-sequence versus prefix comparison uses the separate sampling protocol in Section~\ref{app:prefix-comparison}.

\subsection{Training and attention timing}
\label{app:training-measurements}

Complete-step training measurements include the forward and backward passes, distributed communication, and optimizer updates. The training summary reports median step time after the first 20 steps and takes per-GPU throughput from Megatron's logs. The main paper reports that throughput in tokens/s/GPU and peak GPU memory in GiB for each complete training configuration.

Attention microbenchmarks use one B200 with synthetic BF16 inputs matching the training microbatch sizes: batch four at 4K and batch one at 32K, with 32 query heads, eight KV heads, and head dimension 128. The reported protocol uses six rounds of 20 CUDA-graph replays and alternates configuration order. Reported attention latencies are medians. The BF16 attention timing uses cuDNN, whereas the complete-training BF16 recipe permits Transformer Engine's automatic backend selection.

Forward timing includes input preparation and saved training state; backward timing includes its preprocessing. Forward, backward, and whole attention are measured separately, so their reported medians need not add. Preparation, completed-block attention, and boundary-plus-merge stage timings are also independent measurements. These attention timings exclude model projections, MLPs, and optimizer work; complete-training throughput includes them.

\subsection{Single-request serving throughput}
\label{app:serving-measurements}

Serving uses the released \texttt{Qwen/Qwen3-4B-Base} and \texttt{Qwen/Qwen3-14B-Base} checkpoints on one B200 with vLLM 0.31.0. Weights and projections remain BF16. The compared configurations are BF16 attention with BF16 K/V storage, BF16 attention with FP8 K/V storage, FP8 attention with FP8 K/V storage, and our MXFP8 implementation. The FP8 configurations use fixed per-tensor scales of 1.0, matching the uncalibrated configuration discussed by vLLM \citep{kubler2026fp8kvcache}. Our cache uses data-dependent block scales restricted to observed tokens.

Measurements use CUDA graphs and one request at a time, with repeated-sentence prompts at 4K, 8K, 16K, and 32K context and 128 generated tokens. Each reported throughput averages the medians of five requests in each of two rounds, with configuration order reversed between rounds. Differences below approximately 3\% are comparable to observed run-to-run variation. The measurements compare complete inference implementations under this single-request workload.

Our integration uses single-token decoding, unchunked prefill, and disabled prefix caching. Section~\ref{app:cache} describes the cache representation and update order.

\subsection{Paired prefill NLL}
\label{app:prefill-nll}

For each released model, prefill fidelity is evaluated on 16 C4 sequences of 8,192 tokens. Each sequence contributes 8,191 next-token labels, for 131,056 scored labels per model and configuration. For sequence $s$ and method $m$, the mean negative log-likelihood is

\begin{equation}
\label{eq:supp-nll}
\ell_{s,m}
=-\frac{1}{8191}\sum_{t=0}^{8190}
\log p_m(x_{t+1}\mid x_0,\ldots,x_t).
\end{equation}

The comparison with FP8 attention and FP8 K/V storage is paired by sequence. With $\Delta_s=\ell_{s,\mathrm{ours}}-\ell_{s,\mathrm{FP8}}$, the reported difference and its standard error are

\begin{equation}
\label{eq:supp-nll-paired}
\begin{aligned}
\overline\Delta&=\frac{1}{16}\sum_{s=1}^{16}\Delta_s,\\
\operatorname{SE}(\overline\Delta)
&=\sqrt{\frac{\sum_{s=1}^{16}(\Delta_s-\overline\Delta)^2}{16\cdot15}}.
\end{aligned}
\end{equation}

The differences are $-0.0035\pm0.0012$ nats on 4B and $-0.0016\pm0.0004$ nats on 14B, where uncertainty is one paired standard error across sequences. NLL is comparable to BF16 attention with FP8 K/V storage. This comparison concerns prefill on the stated sample; it does not measure incremental value-rescaling error during decode.
\end{document}